\documentclass[lettersize,journal]{IEEEtran}
\usepackage{amsmath,amsfonts}
\usepackage{algorithmic}
\usepackage{algorithm}
\usepackage{array}
\usepackage[caption=false,font=normalsize,labelfont=sf,textfont=sf]{subfig}
\usepackage{graphics}
\usepackage{adjustbox}
\usepackage{textcomp}
\usepackage{stfloats}
\usepackage{url}
\usepackage{verbatim}
\usepackage{graphicx}
\usepackage{lscape}
\usepackage[backend=bibtex,style=ieee]{biblatex}
\usepackage[figuresright]{rotating}
\UseRawInputEncoding
\begin{document}

\title{Deep Neural Networks in Video Human Action Recognition: A Review}

\author{Zihan Wang, Yifan Zheng, Yang Yang, Moyan Zhang, Yujun Li
        % <-this % stops a space
\thanks{This paper was produced by the IEEE Publication Technology Group. They are in Piscataway, NJ.}% <-this % stops a space
\thanks{}}

\markboth{Journal of \LaTeX\ Class Files,~Vol.~14, No.~8, August~2023}%
{Shell \MakeLowercase{\textit{et al.}}: A Sample Article Using IEEEtran.cls for IEEE Journals}


\maketitle
\begin{abstract}
Currently, video behavior recognition is one of the most foundational tasks of computer vision. The 2D neural networks of deep learning are built for recognizing pixel-level information such as images with RGB, RGB-D, or optical flow formats, with the current increasingly wide usage of surveillance video and more tasks related to human action recognition. There are increasing tasks requiring temporal information for frames dependency analysis. The researchers have widely studied video-based recognition rather than image-based(pixel-based) only to extract more informative elements from geometry tasks. Our current related research addresses multiple novel proposed research works and compares their advantages and disadvantages between the derived deep learning frameworks rather than machine learning frameworks. The comparison happened between existing frameworks and datasets, which are video format data only. Due to the specific properties of human actions and the increasingly wide usage of deep neural networks, we collected all research works within the last three years between 2020 to 2022. In our article, the performance of deep neural networks surpassed most of the techniques in the feature learning and extraction tasks, especially video action recognition.  
\end{abstract}


\section{Introduction}
\IEEEPARstart{V}{ideo} Human Behavior Recognition is widely utilized in multiple research areas, including scene graph generation, Visual Question Answering on multi-modality data, and attention-based feature generation. Due to the previously finished research, deep learning methodology has achieved outstanding identification performance and prediction tasks on video-based tasks. There are some previously investigated studies \cite{1} working on the improvement of human action classification tasks by generating probability through multiple streams or the combination of models. Our research work investigations are based on the modality data, including the skeleton, RGB, RGB+D, optical flow, depth map, etc. The application scenarios are comprehensive, including driver abnormal behavior identification, scene graph relationship analysis, and pedestrian re-identification. Our investigated frameworks involve individual neural network architectures and the combination of multiple neural networks, such as two-stream feature map analysis and multi-stream analysis tasks. Some work \cite{2} is based on modality data such as audio, text, video, and images when data fusion is utilized for human action recognition. Multiple models are proposed to improve the classification accuracy of computer vision tasks, especially in studies involving human-human interaction and human-object interaction.

This survey is structured as follows: Section II provided the investigation of related works of existing surveys and the latest usage of deep learning on video-based recognition tasks; Section III introduced the datasets involved in current tasks and their descriptions; Section IV provides the popular deep-learning based methodologies in the tasks of human behavior recognition works; Section V involved most of the application scenarios of action recognition in research perspective; Section VI introduced the frameworks of deep learning methods(neural networks) and their structures, as well as the comparison of frameworks; Section VII provided an overview of existing temporal-based block for solving the problems of time-sensitive tasks on human action recognition; Section VIII illustrated the 3D convolutional neural networks with multi-stream and double-stream structures; Section IX provided the information of existing obstacles need to be solved and waiting to be observed requiring further investigation works; Finally, we summarized all the related works mentioned in our articles and provided the futures of deep learning-based methodology in most of application scenarios. Also, this is the cited version of the conference paper "A survey of video human behavior recognition methodologies in the perspective of spatial-temporal".

\subsection{Research Overview}

After we analyzed multiple kinds of research from previous years, including action identification from some popular databases, the bellowing word cloud in Figure 1 reflects the relationships of these topics, composed of numerous popular topics mentioned in current research areas. The existing methodologies have implemented an excellent performance on the behavior classification and prediction tasks with 100\% accuracy, surpassing most current methods. In contrast, some problems \cite{3} are also waiting to be tackled.

\begin{figure}[h]
\centering
  \includegraphics[width=6cm, height=3.5cm]{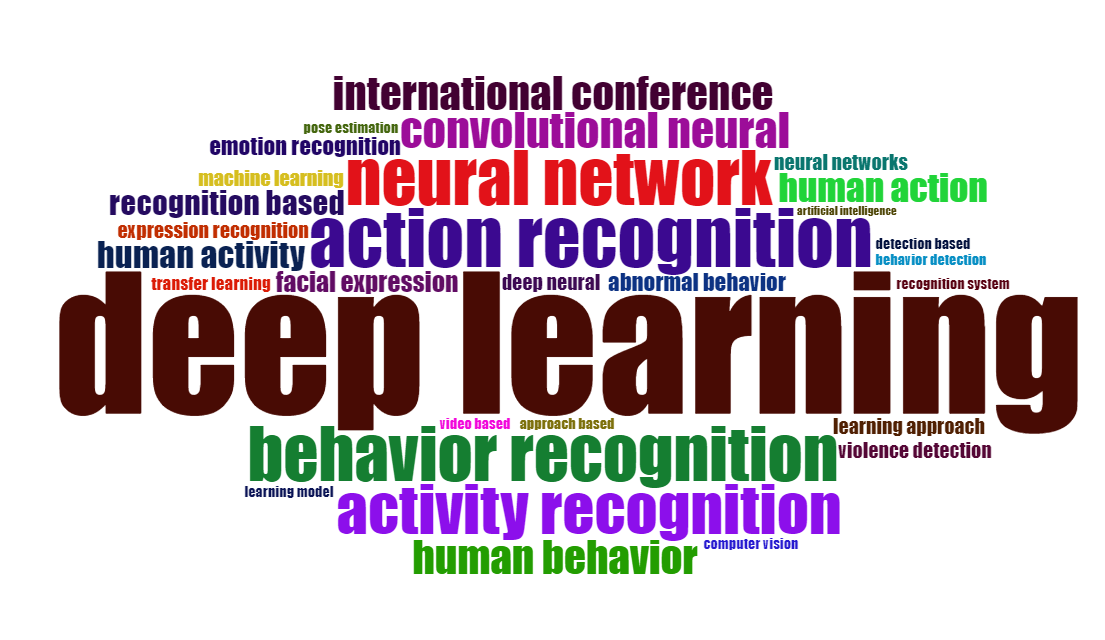}
  \caption{Word cloud - action recognition in deep learning.}
  \label{Figure}
\end{figure}

Such as only a small amount of data could be analyzed due to the large size of video data segments, or the complete video segments need to be sliced batch by batch due to the computation requirements being various according to the type of tasks. Also, video recognition tasks require a large amount of CPU/GPU running costs, while deep learning works require time-dependency analysis. This type of problem still needs to be reconsidered and improved. The above summary graph illustrates the significant topics mentioned in the action detection works, followed by detailed descriptions. Compared with most deep learning methodologies, Convolution Neural Network(CNN) involves the most common usage, and many derived methodologies are manipulated rather than any other deep learning methodologies, such as Graph Convolutional Network(GCN), the skeleton-based and modality data have occupied most usages of the existing datasets, and both of them work as separate branches of video-based recognition tasks. 

The significant contributions of our research works are summarized as bellowing; our work mainly focuses on video recognition tasks with the usage of deep neural networks, especially in deep learning-based methodologies. Due to the wide usage of video-type data, such as surveillance video, TV series, video streaming, or any other scenarios of video segments in real-life situations. The area of deep learning has received great attention and awareness from lots of research works. We summarized some previously posed surveys and existing methodologies in our review to provide an overview of current research situations. 

\section{Related Work and History}

We investigated multiple previously completed research surveys(from 2013 to 2022) of human behavior recognition, including multiple tasks, such as human detection, activity analysis, and mathematics methodologies. Since color information does not play a critical role in most of the video related tasks, because there are multiple categories of data format, such as optical flow and gray-scale are widely used in most video recognition tasks, and the usage of a single type of these data are way less than the modality data. The work \cite{5} in 2013 summarized both spatial and temporal features of recognition methods on multiple types of video images, such as silhouettes, feature representations, and RGB/RGB-D data. Still, it lacked descriptions of recognition methods. For example, the survey \cite{4} from 2015 summarized the crowd(group) behavior recognition methodology with fixed direction movement and discussed the granularity of video content where each pedestrian is treated as a granularity. The survey also analyzed the automatic group people analysis methods called automatic recognition tasks in stream data. The methodologies mentioned in the above survey considered both temporal and spatial features extracted from video pixels and time-sensitive dependency relationships but the deep learning methods are missing from the above work. The temporal blocks for the frame dependency tasks are also considered during \cite{li2020tea} this work for the temporal and spatial behavior recognition in the process of frame aggregation steps. 

The survey work \cite{157} from 2017 went through the vision tasks for human behavior recognition as a survey work describing the human-environment interaction tasks through utilizing the deep neural networks through stacked spatial and temporal blocks for the fusion of motion detection. The comprehensive review \cite{159} from 2019 summarized vision-based tasks for the usage of deep learning-based methodologies with multiple series of datasets in the video-based tasks in the usage of deep learning methods. The latest work \cite{158} from 2022 provided the data/datasets modality which is widely utilized in action recognition tasks, and also compared the accuracy of multiple measurement methodologies in multi-modality fusion methodologies. 

The last work \cite{7} also illustrated the overview of deep learning methods of human behavior recognition which compares the existing frameworks and their structures with labeled accuracy, but it lacked the theoretical description and the technique comparison of methodologies. Our work summarized existing works datasets, backbone frameworks, and deep learning-based works from both pixel and time-series levels. We will further describe the directions of video analysis task development on action recognition from human-human and human-object tasks.

\begin{figure*}
  \centering
  \includegraphics[width=18.5cm, height=5.5cm]{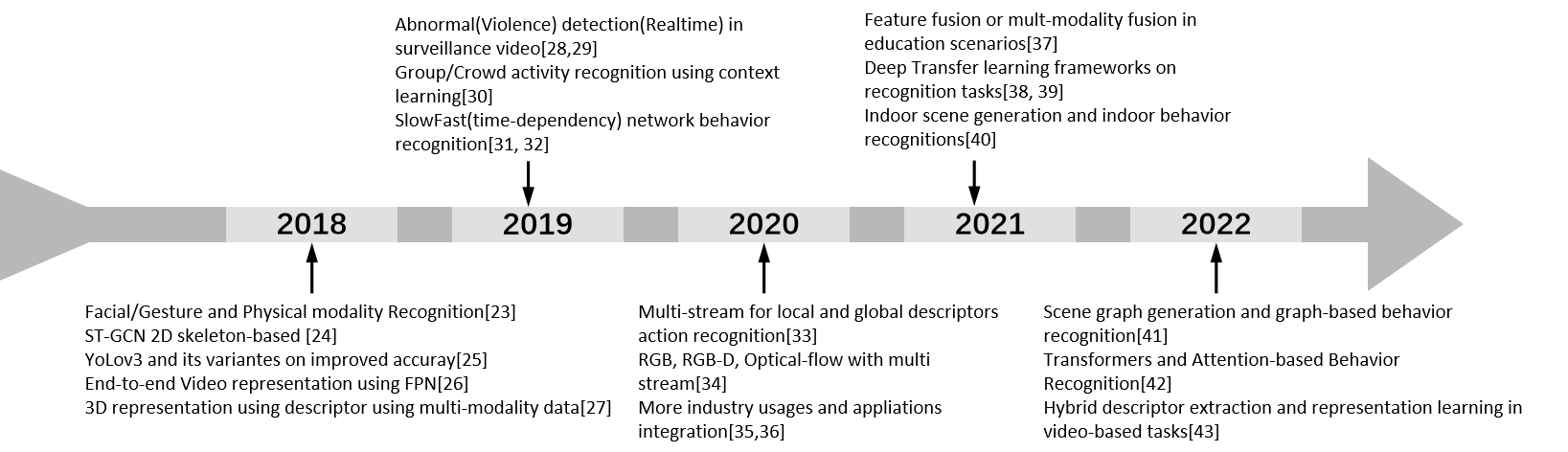}
  \caption{Research development trend of video human action identification.}
  \label{fig:boat1}
\end{figure*}

\vspace{-1em}
\subsection{Timeline of Related Researches}

We also analyzed some of the most famous works, such as ST-GCN or 2s-AGCN, which could represent the trend and novel status of previous video behavior recognition works. As shown in the timeline of Figure 2, from 2018 to 2022, the revolution of this type of task has changed a lot. The task-solving methodology moves from pixel-based static analysis to real-time stream analysis, from machine learning methods to more deep learning methods, and from individual human-based tasks to human-environment interaction tasks. Also, with the popularity of vision transformers, attention is involved in action recognition tasks. In research works that are related to LSTM \cite{13, 14}, they utilize Recurrent Neural Networks(RNN) for multiple stream action recognition and manipulate gated RNN to detect the time-series dependent tasks. Also, the double stream networks are famous for solving human action recognition tasks through utilizing attention-based LSTM in research work with recognition score fusion \cite{15}.

Meanwhile, most deep learning-based methods have been proposed during the last three years \cite{16,17,18,19,20,21,22}, which mention that both supervised and unsupervised deep learning methodologies could solve the recognition tasks according to various data formats and labeling methods. The above timeline shown as Figure 2 \cite{23,24,25,26,27,28,29,30,31,32,33,34,35,36,37,38,39,40,41,42,43} illustrates the methodology revolution process from 2018 to the most recent. The timeline says that more combinations of methodologies are developed rather than individual model usage. Meanwhile, the recognition tasks required more modality data in the steps of data fusion and feature extraction rather than the particular data type with the requirement of recent year works. 

There are also plenty of works utilizing the principle of ST-GCN to accomplish the recognition tasks by graph convolutional networks since 2018. In addition, we found an increasing number of works focusing on human motion recognition tasks, including gait, gestures, and skeleton behavior recognition \cite{44,45, tu2022joint}. Until recent years, some works have also mentioned unsupervised and supervised deep learning methods for behavior recognition tasks.

\section{Datasets Overview}
The data resources \cite{46,47,48,49,50,51,52,53,54,55} of action recognition datasets mostly came from institutions or websites containing many video segments. The scale of the datasets (YouTube, surveillance systems, universities, etc.) is also growing over time, reflecting the number of videos or tags.

\vspace{-1em}
\subsection{Expanding History of Datasets}
\begin{table*}[!hb]\fontsize{5.5}{6.2}\selectfont
\caption[font=12pt]{Datasets Overview}
\hspace{-3.5em}
\begin{tabular}[font=8pt]{l|c|c|c|c|c|l|l|l}
\hline
Dataset Name                                                                      & \begin{tabular}[c]{@{}l@{}}Publish\\ Year\end{tabular} & \begin{tabular}[c]{@{}l@{}}Modality\end{tabular} & \begin{tabular}[c]{@{}l@{}} Group or \\Individual \end{tabular} & \begin{tabular}[c]{@{}l@{}}Video\\ Num\end{tabular}         & \begin{tabular}[c]{@{}l@{}}Action \\Num\end{tabular} & \begin{tabular}[c]{@{}l@{}}Comparison \\Measurement\end{tabular}  & Mentioned in Recent Research                                                                                                                                                                        \\ \hline
\begin{tabular}[c]{@{}l@{}}Violent Flows - \\ crowd violence \cite{46}\end{tabular} & 2012                                                   &  RGB, Text                                                           & Group             & 246                &  2           & ACC, SD, SE, AUC    & \begin{tabular}[c]{@{}l@{}}Violence Detection by MobileNetV2 and LSTM \cite{56}\\ Temporal Cross Fusion Network for violence detection \cite{57}\end{tabular}                                                  \\ \hline
UCF50                                                                             & 2012                                                   & RGB                                                                & Individual        & 150                & 50       & ACC, MAP, SD    & \begin{tabular}[c]{@{}l@{}}3DCNN for Human Activity Classification 
 \cite{58}\\ Dual-stream Deep feature extraction for action classification \cite{59}\end{tabular}  \\ \hline
UCF101                                                                            & 2012                                                   & RGB                                                                 & Individual        & 13320              & 101     & ACC, MAP, SD     & \begin{tabular}[c]{@{}l@{}} Shuffle-invariant network for action recognition in videos\cite{175}\end{tabular}                                        \\ 
\hline

NTU-RGB + D 60\cite{47}                                                               & 2016                                                   &   RGB,Skeleton,Depth,etc                                                         & Individual        & 56880              & 60      & X-Sub, X-View, X-Set     & \begin{tabular}[c]{@{}l@{}}Dual stream cross-modality data fusion transformer for action recognition \cite{62}\\ Multi-modality knowledge embedded GCN for behavior recognition \cite{63}\end{tabular}         \\ \hline

\begin{tabular}[c]{@{}l@{}}Kinetics-400\end{tabular}                                                                       & 2017                                                   &    RGB                                                             & Individual        & \textgreater{}306k & 400      & ACC, Top-1, Top-5    & Temporal action detection with video representation learning \cite{66}                                                                                                                                     \\ \hline
\begin{tabular}[c]{@{}l@{}}Something Something\\v2 (v1 is outdated)\end{tabular}                                                           & 2018                                                   &  RGB                                                           & Individual        & \textgreater{}220k & 174    & ACC, Top-1, Top-5    & Frozen clip models are efficient video learners\cite{176}                                                                                                                                                                           \\ \hline
Kinetics-600                                                                      & 2018                                                   &   RGB                                                              & Individual        & 480k               & 600    & ACC, Top-1, Top-5      & \begin{tabular}[c]{@{}l@{}}Adaptive vision transformer for video sampling \cite{67}\\ Violence detection and Pose Estimation with neural networks \cite{68}\end{tabular}              \\ \hline
\begin{tabular}[c]{@{}l@{}}Atomic Visual\\ Actions(AVA)\cite{49}\end{tabular}                                                  & 2018                                                   &  RGB                                                           & Group             & 430                & 80     & NPMI, mAP, IoU      & \begin{tabular}[c]{@{}l@{}}Long Short term context aware Atomic-Action detection \cite{69}\\ Attention-aware Human-based behavior recognition \cite{70}\end{tabular}                                         \\ \hline
Moments In Time \cite{50}                                                           & 2018                                                   &   RGB, Auditory                                                          & Individual        & 1 million        & 339     & ACC, Top-1, Top-5      & Actionformer: Localizing moments of actions with transformers\cite{177}\\ \hline
UCF-Crime                                                 & 2018                                                   &  RGB                                                           & Group             & 1900               & 13         & AUC, RMSE, F1 score     & \begin{tabular}[c]{@{}l@{}}Crime Detection in Surveillance Systems \cite{71}\\ Surveillance Video Anomaly Detection in 3D convolution neural networks \cite{72}\end{tabular}                        \\ \hline
Hr-Crime        & 2021                                                   &   RGB, Text           & Group            &  1900              &  13        & ROC, AUROC      & \begin{tabular}[c]{@{}l@{}}Spatial-temporal in surveillance video recognition \cite{boekhoudt2022spatial}\\ Multi-pedestrains anomaly behaviour detection\cite{abdullah2023multi} \end{tabular}                        \\ \hline

NTU-RGB + D120 \cite{51}                                                            & 2019                                                   &  RGB, Skeleton, etc                                                          & Individual        & \textgreater{}114k & 120     &   X-Sub, X-View, X-Set      & \begin{tabular}[c]{@{}l@{}}Adaptive Graph Convolutional Networks for skeleton-based recognition \cite{75}\\ IGFormer: Interaction Graph transformer for skeleton-based recognition \cite{76}\end{tabular} \\ \hline
Kinetics-700                                                                      & 2019                                                   &  RGB                                                               & Individual        & 650k               & 700     & ACC, Top-1, Top-5    & \begin{tabular}[c]{@{}l@{}}Time-consistent feature extraction on Action Recognition tasks \cite{77}\\ New Dataset proposed called MetaVD \cite{78}\end{tabular}                                              \\ \hline
\begin{tabular}[c]{@{}l@{}}Multi-moments \\In Time \cite{52} \end{tabular}                                                    & 2019                                                   &   RGB                                                          & Individual        & 1.02 million       & 292    & Top-1, Top-5, mAP, AUC     & \begin{tabular}[c]{@{}l@{}}New Dataset Proposed called Multisports \cite{79}\\ 3D spatial-temporal glance and combine module based on RGB DATA \cite{80}\end{tabular}                                        \\ \hline
Countix Data Series \cite{zhang2021repetitive}                                                              & 2021                                                   &  RGB                                                           & Individual        & 8757               & 100    & MAE, OBO     & Effective Approach to Repetitive Behavior Classification based on Siamese Network \cite{173}                                                                                                                                                                                        \\ \hline
Home Action Genome \cite{55}                                                        & 2021                     & RGB, Infrared, etc                                                             & Individual        & 5700               & 70     & mAP, AUC     & Scene graph generation with visual relationship detection  \cite{81}                                                                                   \\ \hline

HDM05 \cite{hdm05}                                                          & 2007                                                 &  Optical, Skeleton, RGB                                                         & Individual         &  1500              & 100 & ACC, Std.         &  Multi-radii circular signature based feature descriptor for hand gesture recognition \cite{165}                                                   \\ \hline

MSR-Action3D \cite{msrac}                                                        & 2010                                                 & Skeleton, Depth                                                           & Individual        &     567           &   20    & ACC       &  \begin{tabular}[c]{@{}l@{}} Deep Learning-Based Human Action Recognition with \\Key-Frames Sampling Using Ranking Methods\cite{166} \end{tabular}                                                     \\ \hline

MSRC12 Gesture \cite{msrc}                                                      &  2012                             &  RGB,Skeleton,Depth                                                     & Individual           &  594              &  12      & ACC        &  Adaptive hough transform for hand gesture recognition\cite{167}                                                   \\ \hline

PKU-MMD \cite{pkummd}                                                        &  2017                            &   RGB,Skeleton, Infrared                                
               &  Both       &  1076                &    51     & F1, AP, mAP, 2D-AP        &  Contrastive 3D Human Skeleton Action Representation Learning\cite{168}                                                   \\ \hline
(Multi-)Thumos\cite{multi}                                                        &   2015                           &  RGB,Text\footnote[1]                                                 &  Both       &  400               &  65    & Detection-score, mAP, AP         &  GCAN: Graph-based Class-level Attention Network for Long-term Action Detection \cite{169}                                                   \\ \hline
Charades\cite{charade}                                                        &   2016                              &  RGB                                                   &  Individual        & 9,848                &    157   & mAP, AP        &  Transferable Network for Zero-Shot Temporal Activity Detection \cite{174}                                                  \\ \hline
Charades-STA\cite{sta}                                                        &   2017                                               &  RGB, Text\footnote[1]                                                          & Individual        &  9,848               &  157      & IoU, Recall       & Moment is important: language-based video moment retrieval via adversarial learning\cite{171}                                                     \\ \hline

TVSeries\cite{tv}                                                        &   2016                                               &  RGB, Text\footnote[1]                                     & Both         &  27              &    30     & mAP, mcAP, cAP       &   Information Elevation Network for Online Action Detection and Anticipation \cite{172}                                                  \\ \hline
XD-Violence\cite{wu2020not}                                                        &   2020                                              &  RGB, Audio                                      & Both         &  4754              &    6     & AP, AUC, mAP, ACC, etc         &   Self-supervised learning for anomaly human behavior detection\cite{panariello2023consistency}                                            \\ \hline
\end{tabular}
\vspace{1em}
\\
{NOTE: $^1$ represents the labels, subtitles, sentences, and metadata where are existing in the provided datasets}\\
\end{table*}

The datasets published earlier in 2012 include Violent
Flows-crowd violence \cite{46} and UCF50, UCF101, from the
expansion and development of datasets. Data sources for these
three datasets are coming from YouTube websites. Violent
Flows-crowd violence is a database of real crowd violence
video clips containing 246 videos with an average video
duration of 3.6s. The number of action categories in UCF50
and UCF101 is represented as the ending number of the dataset's
name with 50 and 101 classes respectively. The main difference between these two datasets is the
number of labels where UCF50 extended from the UCF11
and UCF101 extended from UCF50. However, the metadata
of both are kept consistent, as each category is divided into 25 
groups with more than four videos in each group.

There were also some shortcomings for both action identification datasets, 
which included few labels and non-real
environment recordings. And the videos are uploaded by
real users of video websites. The video frame rate in UCF
is 25 FPS, the resolution is 320px${\times}$240px with the video
format as .avi, and the average video clip duration is 7.21 seconds.
The naming convention is vX\_gY\_cZ.avi, where X represents
the category, Y represents the group, and Z Indicates the video
number. Another scenario-specific dataset called UCF-Crime was proposed in 2018 and later,  mainly composed of real surveillance videos aimed at abnormal behavior datasets,
including 13 odd actions(out of threshold defined in
normal constraints) and 1900 relatively long videos. Among
them, 1610 videos of them are training sets, and 290 videos of them are test
sets.

NTU RGB+D 60\cite{47} was developed followed by NTU RGB+D 120 in 2016. This dataset contains
60 types of actions and a total of 56,880 samples, where 40 types are daily actions, 9 are
healthcare-related, and 11 are interactions
between humans. The dataset uses three cameras for
few-shot videos coming from different angles to collect depth
information, 3D bone information, RGB video, and infrared
sequences. The resolution of RGB video is 1920px ${\times}$ 1080px, and the depth map and infrared video are both 512px ${\times}$ 424px, respectively. The 3D bone data is the 3D coordinates with 25 joint points. The training and test sets are labeled by person id and camera type.

The Kinetics-400 dataset was proposed in 2017. After that, Kinetics-600 and Kinetics-700 were proposed in 2018 and 2019. The dataset name is followed by the number of labels the video contains. Each actions category of Kinetics-400 has at least 400 video clips, and the length of the video clip averaged 10 seconds for each segment. All actions can be divided into 38 categories by different granularity. Expansion happened between the kinetics-600 dataset and the kinetics-400 dataset, including the number of labeled types and video numbers in each category. The same extension happened between Kinetics-700 and Kinetics-600 datasets as well.

Moments in Time (Mit) \cite{50} was proposed in 2018, including 339 labels with more than 1,000,000 videos. The author of the datasets mentioned that the available human working memory length in neurons is 3 seconds on average, and video segments with 3 seconds contained enough information for neural analysis with the usage of deep neural networks. The unique feature of design in Mit is to
make models learn the conceptions of actions with solid generalization ability. If a moment can also be heard in the video (e.g., ”clapping” in the background), then we include it with the sound rather than pixels only. 

The datasets proposed in recent years also include Countix
Data (2020 CVPR) and Home Action Genome (2021 CVPR). The Countix \cite{53} Data is a companion dataset of a new network model named RepNet. To increase the semantic
diversity of the previous duplicate dataset (QUVA, a dataset
including multiple sports activities) and expand the scale of
the Countix dataset. It is a real-world duplication of a wild-collected video dataset. A wide range of semantic settings is covered, such as camera/object motion, and the changes in the speed of repetitive actions. Countix data consists of high-frequency activities in Kinetic.
Thus it is a subset of Kinetic but 90 times larger than the
previously posed QUVA dataset.

Home Action Genome (HOMAGE) [55] is a large-scale
dataset proposed in 2021 CVPR, which has multi-view and
multi-modality data for facilitating learning from multi-view
and multi-modal data. It contains 1752 synchronized sequences
and 5700 videos divided into 75 activities and 453
atomic actions. Each series has one high-level activity category.
The property of atomic movement is that the duration
of each action series is usually short. 

We are not going to include all the details of datasets due to large amounts of information and their applications. The remaining detailed information could be checked in the references of datasets and their application scenarios according to the volumes, modalities, number of actions, and pixel sizes of each video/image segment.

\section{DEEP LEARNING-BASED METHODOLOGIES}

This section illustrated some procedures on specific datasets, such as contrastive learning for both supervised and unsupervised scenarios. The critical drawback of supervised learning has been mentioned where the labeling data requires plenty of human labeling work and the large involvement of disciplines, which made the recognition work difficult and time-consuming, especially when two expressions or gestures have high similarity \cite{82,83} between frames by using weakly supervised learning. As mentioned in the above research, weakly supervised learning includes three critical properties incomplete, inexact, and inaccurate supervision. These three steps are accomplished before the classification tasks are finished. 

\vspace{-1em}
\subsection{Contrastive Learning}

Contrastive learning could be utilized in both supervised and unsupervised methodologies. This method is proposed under the circumstance of large batch datasets and more training steps are required when it worked as unsupervised learning for human action recognition. The initial proposal of this framework is to solve the visual representation in 2020 \cite{84}. The contrastive domain adaption has been proposed in the research work to solve the additional temporal dimension on video analysis \cite{85}. Contrastive learning could also be utilized in anomaly detection in surveillance video \cite{86}. It is to make the samples with the same classes close to each other and make the margin distance larger with different classes, which are intra-class and inter-class. The framework proposed in the research solved the anomaly detection task by capturing discriminative semantic features \cite{86}. They solved the problems when new samples appeared in the video, as the newly inputted samples should not be recognized as a disorder while the recognized behavior is within the threshold of defined normal constraints. Also, the discriminative semantic features are also extracted for fixed patterns and context-aware environment recognition by utilizing contrastive learning when the behavior analysis depends on the human interaction with the environment \cite{86,87}. 

\subsection{Measurement/Evaluation Metrics}
There are multiple types of metrics are utilized for the measurement of video-based behavior recognition. The multiple types of measurements include the Average Precision(AP) and mean Average Precision(mAP), which are widely in the usage of behavior recognition tasks. It depends on which modality is suitable for the behavior recognition tasks, such as the cross-view, cross-subject, and cross-setup are widely utilized in NTU-RGB+D datasets. Whereas, some commonly utilized measurements also include top 1 accuracy or top 5 accuracy or accuracy for the behavior recognition in the specific category. Also, the evaluation of measurement is also depending on the selection of features and the defined classifiers according to the feature generation in supervised methodologies, such as precision, recall, and F1-score which are mostly utilized by other works. \cite{1001}The loss function could also be calculated through the designed loss function or mutual information, such as \cite{infogcn} research work defined the mutual loss function according to the stochastic variables, w.r.t the X and Y are mutual information as shown in equation 1.
\begin{equation}
    R(Z) = I(Z;Y) - \lambda_{1}I(Z;X) - \lambda_{2}I(Z;X|Y),
\end{equation}
The loss function could be designed according to different scenarios, such as reconstruction loss function, cross-entropy function, mean square errors, or the class-wise domain loss between categories and classifications. It varies according to the evaluation methodologies, such as the sensitive loss in \cite{SERNA2022103682}, and L1/L2 loss in \cite{jain2023automated} which represents the distance between facial expressions and detected emotions. There are definitely multiple categories of loss functions that are defined according to the change in situations and behaviors. 

\vspace{-0.5em}
\subsection{3D and 2D Skeleton(Multi-modality) Behavior Recognition}

The 2D and 3D skeleton-based pose estimation has been utilized in multiple scenarios. The original designation of human joints pose estimation is to extract the skeleton data as one of the modality data within the range of multiple types of input images as mentioned above. The pose estimation tasks\cite{178} are still popular during the current research situation. Deep learning-based methodologies include such as Graph Interactive Networks, Graph Attention Networks and other types of graph-based neural networks are the main methods of problem-solving. The proposed paper on datasets on NTU RGB+D and NTU RGB+D 120 is the first work that mentioned the measurement of cross-view and cross-subject according to the action category and the views. Similarly, the Human3.6M and 3DPW with the model SPML also extracted the skeletons in the format of 3D. Both 2D and 3D skeleton works are supported for human behavior recognition works in the 3D space. As one of the modalities in computer vision tasks, deep learning methodologies are focusing on feature extraction to finish the tasks of classification, regression, or predictions. The workflow of skeleton behavior recognitions is basically the end-to-end learning or machine learning pipelines for further behavior predictions or downstream tasks. Some of the works accomplished the embeddings of visualizations or contrastive learning to go through the process of comparisons between multiple behaviors. 

\subsection{Domain Adaptation}
\vspace{-1em}
Utilizing domain adaptations on both source and target datasets has provided a convenient way to recognize the classifications of datasets on human action recognition, from the perspective of from source to target or from target to source. Unsupervised research proposes the novel multiple data source included unsupervised domain adaption \cite{91} for semantic recognition rather than single-source unsupervised domain adaptation(UDA). This research work utilized adversarial domain adaptation by generating the synthesis image and data for pattern recognition and recognized by discriminator in GAN to separate real data and synthesize data. 

The research work \cite{92, 93} from Tencent utilized domain adaption combined with GAN achieved to keep the intra-video content consistent and prevent video content distortion. According to the survey of unsupervised domain adaptation \cite{90}, the research work also illustrated generative adversarial DA during the competition of 2 networks to improve the recognition task accuracy. During the research work \cite{91}, the intrinsic discrimination features are extracted from the source domain, which could be utilized for behavior pattern recognition for the target domain. This conception is also mentioned in research work which works as a feature encoding technology in computer vision tasks \cite{91}. Utilizing transfer learning, such as domain adaptation is to learn the specific patterns from source to target domain which has been verified with outstanding performance on action recognition tasks. 

Another research work described the process of how to apply the source domain information from labeled data onto target unlabeled data for recognizable feature representation and recognition. The proposed architecture \cite{94} utilized a multi-headed(two-headed) mechanism network and multi-branch structure to minimize the entire recognition loss of target domain recognition. In the research work \cite{94}, It manipulates the video domain adaptation onto large-scale video datasets where the domain adaptation neural network is considered as DANN rather than DCNN in research work \cite{95} as well. There are two critical conceptions mentioned in the research works \cite{96, 97}. One is domain shift operation which is to get the recognized information from one domain to another. Another conception is an adversarial adaptation which is mentioned in lots of combinations of GAN(Generative Adversarial Network) and unsupervised domain adaptation.

In the above research works, the source feature vectors and target feature vectors constructed the Pixel Correlation Discrepancy(PCD) and Pixel Correlation Matrix(PCM) to calculate the relationship between feature dimensions \cite{97}. As above, one of the critical steps is to understand the correlation matrices for the source domain and target domain separately to determine the discrepancy of pixel correlations. Then this type of method, as mentioned in \cite{97}, is to input both feature vectors as PCM into Gradient Reverse Layers(GRL) and generate the PCD matrix. The target is to apply the domain adaptation operator from source to target and from target to source. And the accuracy of the first operation on PCD is 83.8\%, while the second operation on PCM is 56.1\%. 

\section{Application Scenarios And Methodologies}

With the applications of video-based behavior recognition, the research focuses on dementia-suffering elders as the specific type of behaviors are frequently disordered and repeatable for investigating odd action prediction and prevention \cite{98}. And the convolutional neural network also achieved excellent feature extraction performance in several scenarios. Same as above, the most popular YOLO series of applications is to accurately recognize human actions through a bounding box and has gained significant recognition percentages in several scenarios. 

\vspace{-1em}
\subsection{Scene Graphs}
In the beginning, the scene graph mentioned the relationships of humans and objects according to locations, events, time series, etc \cite{99}. Scene graph refers that putting objects in a stereoscopic environment, such as in front of, behind of, on the side of, and other types which could mention the geographical information of existing objects. The events referred to the interaction between humans or sports, like picking up the cup, holding objects, and carrying stuff as shown in Figure 3. The relationship between humans and objects is composed of a Directed Acyclic Graph(DAG), representing the relationship between humans and the environment.

\begin{figure}[ht]
    \centering
    \subfloat{\includegraphics[width=0.22\textwidth]{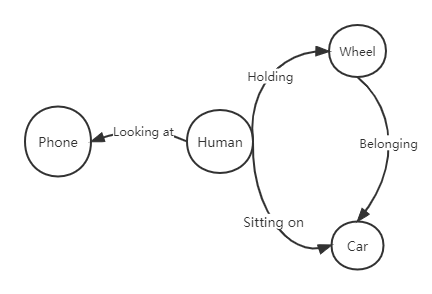}}
    \subfloat{\includegraphics[width=0.22\textwidth]{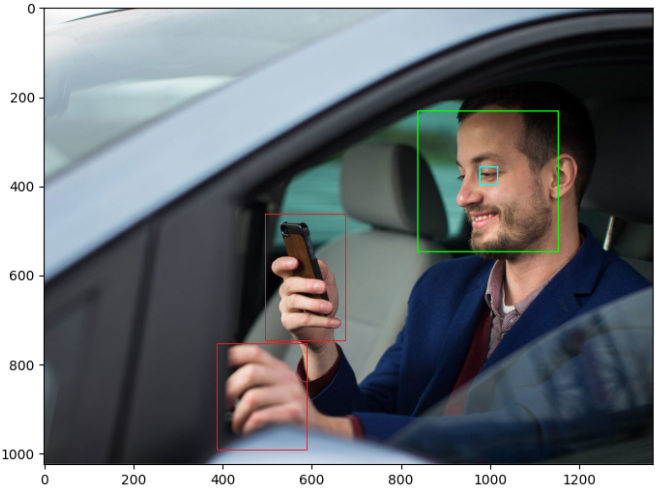}}
    \caption{Human-scene interaction scenarios.}
    \label{fig:foobar}
\end{figure}

The time series is utilized for moving objects rather than static events, especially in video content analysis when mentioning the relationships of human-human and human objects using scene graphs. But the limitation of the scene graph is obvious when it does not consider synchronized human behavior, for example, holding the cups while listening to the smartphone. The synchronized behavior patterns shown above are more general than relationships between humans and objects, represented as two synchronized graphs with common vertices which are utilized for behavior recognition. 

\vspace{-1em}
\begin{figure*}
    \centering
    \includegraphics[scale=0.7]{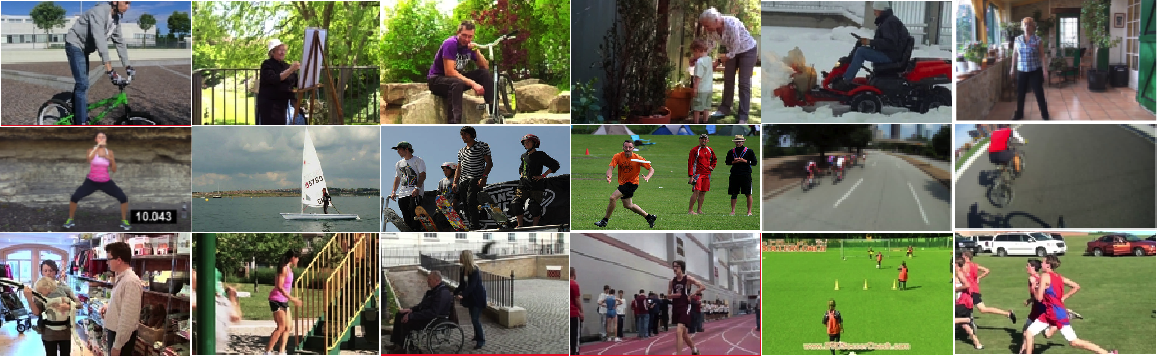}
    \caption{A group of behaviors that are available for the deep learning-based behavior recognition tasks.}
\end{figure*}

\subsection{Extracted Behavior Recognition Descriptor}
The descriptor of behavior recognition means the features of learning and identifications when the type of video sequences includes both discrete and continuous human parts descriptors. The convolution neural network is the primary framework feature extraction procedure within the deep learning methodology. In the research works \cite{100} they manipulated GRU to generate human captions based on video sequence frames, which could implement the recognition tasks on time series. As shown in Figure 4, which are some behaviors coming from multiple datasets are grouped together for the behavior classification tasks.

\vspace{-1em}

\subsection{Multi-modality Behavior Recognition}
The multiple data modalities provide the context of video human action recognition. The results of the fusion score provide the classification results of video-based motion recognition. As mentioned in most research works, data fusion includes three fusion methods: late fusion, early fusion, and hybrid fusion. The data modality will decide the data processing methods and extract multiple types of features. The hybrid architectures of data fusion gave the final results of classification possibility. The modality data also include various input formats, such as optical flow and infrared data, RGB, and RGB+D data. From the information level, the audio, text, caption, and subtitle of the video are also great sources of motion recognition tasks. By fusing the multiple data types into one fusion score, we illustrated both motion and static behaviors that could be recognized, which provided improved recognition results. 

The previous work summarized the skeleton behavior recognition\cite{156} which could be traced to early 2016. As for the skeleton type of data segments, the features above are bones and joints, which could represent the human pose estimation percentage. Several types of skeleton-based poses could be analyzed when their scenario is under temporal consequences \cite{101}. Skeleton-based methodology with 3D human pose estimation is a popular topic in several recent research works. The rotation-based skeleton poses analysis using lie algebra based on time series with spatial information to represent the angles and turnings of human parts \cite{101}. Both ST-AGCN(spatial-temporal adaptive graph convolutional network)/T-AGCN(temporal adaptive graph convolution network) recognize the skeleton-based human action sequences \cite{99} by using the fundamental theories of ST-GCN. The skeleton extraction algorithm is utilized for 3D skeleton-based behavior extraction in the preprocessing steps.
\vspace{-1em}
\subsection{Action Recognition Tasks}

\paragraph{Feature Extraction} Feature engineering plays a critical role in the first stage of human behavior recognition tasks. Due to the specific properties of action data, most video data are composed of multiple formats rather than single formats. Such as both skeleton data and text caption data could be utilized together for modality behavior analysis. The feature extraction step could happen in the first stage or second stage of analysis rather than the pre-processing steps, especially when 2 frames of behavior have high similarity in the feature representations or we call them representation learning. There are multiple works that investigated the methodologies of facial expressions \cite{8} as the emotional representations of humans have high similarity between the previous frame and the current frames. The feature classification \cite{9} could also happen between the classes of pre-defined types, such as some researchers utilizing the Encoder-Decoder architecture for discriminative feature extraction targeting to enlarge the distance between intra-classes and reduce the distance between inter-classes. Also, the current research methods also utilized the Euclidean Distance to measure the similarity of two frames for classification comparison, which are mentioned in multiple research works. The classification accuracy will reach a better performance through the combination of multiple measurements.

\paragraph{Unsupervised Deep Learning} The currently proposed searches such as DBN(Deep Belief Networks) and SVM(Support Vector Machine) research are working on the limited size of data even though both of them have reached the highest identification results. The root cause is most datasets require heavy labeling work to classify the datasets before the tasks started. One of the challenge-solving methods is to reduce the size of datasets by slicing them into a batch format to run the tasks. Another way is to convert the RGB data into optical flow format or gray scales, which made the color information missing from the above works. But the color information occupies the vital rules in computer vision tasks, especially unsupervised deep learning works. The unsupervised methodology is to avoid the large required labeling works through enlarge the distance between different categories and reducing the margins between the same classes.

\paragraph{Attention-based methodology}The attention-based methodologies are also a popular tool for behavior recognition which are categorized into multiple-head and single-head mechanisms with the architecture of query, key, and value. The proposed research works have commonly utilized the soft-attention mechanism for recognition tasks. Hard attention utilizes a stochastic sampling model rather than the weighted parameters in soft attention, where soft attention is more prevalent in human identification tasks. Currently, considerable research focuses on manipulating the attention-based mechanism with skeleton-based behavior recognition tasks. The skeleton data is view-invariant and motion-invariant, which makes the skeleton data suitable for human action recognition tasks \cite{10,11,12}. The combination of attention-based deep neural networks and skeletons became a popular trend in vision-based studies. 

\paragraph{Scene Graph(Indoor Activity) Recognition}

3D stereo pattern recognition has been recognized in several events, including object detection, human pose recognition, and video action analysis. Especially, humans and objects are recognized as vertices of graph nodes while the 'dynamic' relationship between each other is recognized as edges. This type of work utilized encoder-decoder or encoder-decoder-encoder mechanisms for discriminative feature extraction \cite{141}. Also, with the popularity of multi-head attention mechanisms, some frameworks currently utilized transformers as their backbone to analyze video segments by indexing frames. The scene graph generation method is one of the most common graph relationship generation functions, which worked well between objects and humans. The scene graph is considered as the relationship between multiple objects in 3D space, where the video background could be simulated as a 3D environment. The research work of this paper also mentioned how scene graphs analyze video data. The model investigated here is combined with a transformer for generating the scene graph by inputting video data and exploring the relationship between humans, the environment(places), and objects. As the novel proposed research, a scene graph mentioned the relationship between objects in the 3D environment \cite{142}. It built the 3DSSG on top of 3RScan, aiming to extract the 3D geometry and depth information of 3D space. Rather than the 2D scene graph, the involvement of a diagram is more densely connected and informative in the 3D scene graph.  As mentioned in the work \cite{142} of scene graph generation, it allows video analysis of moving objects and human behaviors in a stereoscopic environment.

\paragraph{Online Action Detection(OAD)} Online action detection is also a popular topic with recognizing human behaviors from both spatiotemporal perspectives in the context of the environment. There are only a few works mentioning online action detection rather than online learning which is a widely mentioned methodology in machine learning. The recent years' works are widely combining transformers with online recognition tasks or timely prediction tasks with RNN, such as \cite{160,161,162} research works. As the online detection is to predict the behaviors in time series especially in video streams with the upcoming frames through a defined temporal LSTM encoder-decoder structure. The research works \cite{163,164} mentioned the conception proposal from the research works with the real-time behavior prediction with single frame detection, such as TV series. Online action detection focuses on the tasks of prediction in the usage of backbone feature extraction frameworks. 

\begin{table*}[!hb]\fontsize{5.7}{6}\selectfont
\caption[font=12pt]{Deep Neural Framework Overview}
\hspace{-4em}
\begin{tabular}{l|l|l|c|c|c|c|c|l}
\hline
Neural Network Model        & \begin{tabular}[c]{@{}l@{}}Publish\\ Year\end{tabular} & \begin{tabular}[c]{@{}l@{}}Conf/\\ Journal\end{tabular} & \multicolumn{1}{l|}{Para(M)} & \begin{tabular}[c]{@{}c@{}}Layer\\ Number\end{tabular} & {\begin{tabular}[c]{@{}l@{}}Modality\end{tabular}}  &   \begin{tabular}[c]{@{}c@{}}Fusion\end{tabular} & \begin{tabular}[c]{@{}l@{}}Spatial-\\Temporal\end{tabular} & Description and Comments                                                                                                                                                                                                         \\ \hline
C3D \cite{102}                 & 2014                                                   & CVPR                                                            & 79                             & 8                                                      &    RGB                                            & space-time                          & Yes                                                         & \begin{tabular}[c]{@{}l@{}}It is a spatial and temporal deep 3-dimensional convolutional network \\ which is primarily focusing on video classification and action recognition.\end{tabular}                                        \\ \hline
Fast R-CNN \cite{103}          & 2015                                                   & ICCV                                                            & 3.3                            & -                                                      &    RGB                                            &  image-feature                         & No                                                          & \begin{tabular}[c]{@{}l@{}}A framework with MIT licenses focus on pose estimation, object \\ detection, and classification algorithms. It covers a wide range of detection \\ with more than human-only.\end{tabular}       \\ \hline
Faster R-CNN \cite{104}        & 2015                                                   & NeurIPS                                                         & 3.3-60                         & -                                                      &  RGB                                              &  multi-features                         & No                                                          & \begin{tabular}[c]{@{}l@{}}A framework is faster than above with similar logic for convolutional \\ feature map detection—image processing only rather than video processing.\end{tabular}                                   \\ \hline
SSD \cite{105}                & 2016                                                   & ECCV                                                            & -                              & 11                                                     &   RGB                                             & multi-features                         & Yes                                                         & \begin{tabular}[c]{@{}l@{}}Single-shot detector for real-time image object/human detection. \\ Both  images and videos are suitable for training and testing\end{tabular}                                                     \\ \hline
ResNet152-v2 \cite{106}        & 2016                                                   & ECCV                                                            & 60.34                 & 152                                                    &  RGB                                              & multimodal                         & Yes                                                         & \begin{tabular}[c]{@{}l@{}}Facial and physical detection frameworks include abnormal behavior \\ detection using ResNet and Residual Network.\end{tabular}                                                                       \\ \hline
ResNet101-v2 \cite{106}        & 2016                                                   & CVPR                                                             & 44.65                          & 101                           & RGB                                                 &  image-feature                         & Yes                                                         & Same function as above with a distinct number of neural layer number.                                                                                                                                                              \\ \hline
Inception-v3 \cite{107}        & 2016                                                   & CVPR                                                            & 22.8                           & 42             & RGB                                                &  image-feature                        &  No                                                          & It is an image classification and recognition model trained based on ImageNet.                                                                                                                                                   \\ \hline
Inception-v4 \cite{108}        & 2016                                                   & CVPR                                                            & 42.68                          & 133                                                    &  RGB                                                & image-feature                        & No                                                          & \begin{tabular}[c]{@{}l@{}}As mentioned above, a simplified version of the frameworks are based on Inception blocks\\ and residual blocks for image classification tasks.  \end{tabular}                                                               \\ \hline
Inception-Resnet-v1 \cite{108} & 2016                                                   & CVPR                                                            & 6.977                          & 206                       &  RGB                                                 &   image-feature                     & No                                                          & Same source and similar strcuture as Inception-v4.                                                                                                                                                                               \\ \hline
Inception-Resnet-v2 \cite{108} & 2016                                                   & CVPR                                                            & 55.84                          & 391                          & RGB                                                    &  image-feature                     & No                                                          & Same source and similar strcuture as Inception-v4.                                                                                                                                                                               \\ \hline
I3D \cite{109}                 & 2017                                                   & CVPR                                                            & 25                             & -                                                      & RGB, Optical                                                    & multimodal                      & Yes                                                         & \begin{tabular}[c]{@{}l@{}}A convolutional neural model trained based on Kinetics with human video \\ behavior recognition.\end{tabular}                                                                                       \\ \hline
T3D \cite{110}                 & 2017                                                   & CVPR                                                            & 10.48                   & 121                                                    &   RGB                                                  &  space-time                     & Yes                                                         & \begin{tabular}[c]{@{}l@{}}Transfer learning for video motion recognition framework to analyze both\\  spatial-temporal perspectives of video cube data.\end{tabular}                                                         \\ \hline
Mask R-CNN \cite{111}          & 2017                                                   & IEEE                                                            & 2.16                           & 33                                                     &    RGB, Skeleton                                                  &  decision\footnote[2]-feature                   & Yes                                                         & \begin{tabular}[c]{@{}l@{}}It is an RCNN-based framework with object detections through using bounding boxes. It has been \\ verified what is suitable for computer vision tasks on both images and videos.\end{tabular}               \\ \hline
Xception \cite{112}            & 2017                                                   & CVPR                                                            & 22.8                          & 71                                                     &   RGB                                                   & multimodal                    & No                                                          & \begin{tabular}[c]{@{}l@{}}Image classification framework is trained on top of ImageNet with deep learning methods to\\  learning the features by convolution operators.\end{tabular}                                            \\ \hline
DenseNet-121 \cite{113}       & 2017                                                   & CVPR                                                            & 8.06                           & 121                                                    &  RGB                                                      & image-feature                  & Yes                                                         & \begin{tabular}[c]{@{}l@{}}Densely connected convolutional networks with 121 convolutional layers for image classification.\\  It is verified the video human action recognition in surveillance could be detected .\end{tabular} \\ \hline
DenseNet-169 \cite{113}       & 2017                                                   & CVPR                                                            & 14.31                          & 169                                                    &  RGB                                                      & image-feature                 & Yes                                                         & Same as DenseNet-121.                                                                                                                                                                                                                   \\ \hline
DenseNet-201 \cite{113}       & 2017                                                   & CVPR                                                            & 20.24                          & 201                                                    &  RGB                                                     & image-feature                  & Yes                                                         & Same as DenseNet-121.                                                                                                                                                                                                                   \\ \hline
Res3D \cite{114}              & 2017                                                   & CVPR                                                            & 33.2                           & 18                                                     & RGB, Optical                                                        & time-space                   & Yes                                                         & Video human events recognition tasks framework trained is based on Kinetics datasets.                                                                                                                                            \\ \hline

EfficientNet \cite{116}       & 2019                                                   & IMLS                                                            & 66                             & 813             & RGB                                                          &  order-time-space               & Yes                                                         & \begin{tabular}[c]{@{}l@{}}A framework that could recognize images and videos of human actions is trained based on ImageNet.\end{tabular} \\ \hline
SlowFast \cite{117}            & 2019                                                   & ICCV                                                            & -                              & 50                                                     &   RGB                                                       & modality-decision                 & Yes                                                         & \begin{tabular}[c]{@{}l@{}}A framework for video detection utilized double stream(fast and slow process speed) architecture. \\ It is also could be used for human action recognition.\end{tabular}                             \\ \hline
\end{tabular}
\vspace{1em}
\\
{NOTE: $^2$ represents the decision made through multiple single probability generation streams.}\\
\end{table*}

\section{Deep Neural Network and Comparison}

Many deep learning backbones and deep neural networks \cite{102,103,104,105,106,107,108,109,110,111,112,113,114,115,116,117,118,119,120} have been utilized for action analysis tasks as shown in table II. The deep learning-based method is trying to improve recognition accuracy by stacking more neural layers for feature extractions rather than machine learning algorithms. 

The research work \cite{121} discussed how unsupervised learning works better than supervised methods by manipulating contrastive learning on the vision representation data. As per our verification, most of the vision data could be converted into the format of embedding and compared through contrastive learning. Meanwhile, the research work \cite{122, 123, 124, 125} discussed how deep learning works well in skeleton-based behavior recognition through the common frameworks with GCN and CNN. The double stream of combined score results has been verified its advantages in multiple behavior recognition tasks with double joints stream in research \cite{125}, and the combination of multiple models in deep feature extractions \cite{122}. The multi-modality human behavior recognition has also proven its availability in the increased identification accuracy. The research works \cite{123, 124} investigated the wide usage of modality data and its improved performance on human behavior recognition accuracy on multiple behavior categories. With recent years' research, due to the specification of muti-modality dataset being large-scale and data type varies, the deep models may require more parameters, and an increased number of neural network layers, which made the large-scale model training receive more attention. And the fusion methods are also varying according to the fusion methods which rely on feature extraction and feature representations. The research work \cite{123} discussed how the deep encoder-decoder and multiple deep neural networks could work well together on multi-modality data recognition which performs better than the single-modality data.\\

\vspace{-13pt}
Meanwhile, the time-dependency relationship of consistent behavior analysis also performs a critical role in current research, as human motion is composed of a series of atomic actions in a single behavior. As per the request of current time-related tasks, the temporal attention model accurately addresses the time-based feature extraction through the attention-based models \cite{126, 127}. Then the next step is to optimize the correlation map and loss functions with more converged iterations for the modeling training on the video-based data. The research \cite{128} also investigated the representation of the behavior such as "touch head" need to be represented as both physical connections and non-physical connections relationships. The above research analyzes the connections between local and non-local blocks where multiple individual blocks consider the behaviors. Most deep learning tasks of action representations and abnormal recognition before 2020 were summarized during research work \cite{129}; it outlined most of the machine learning methodologies, some deep learning-based methods, and their application scenarios with the video-based data. But this research work should have noticed how the multi-modality data and multiple branches/streams provided higher accuracy than a single modality in identifying HAR(human action recognition).

\vspace{-1em}
\subsection{Deep Belief Network and Boltzmann Machine(DBM)} 
With the wide usage of feature engineering and representation learning, the multi-modality data could be mapped into low-dimensional spaces, as mentioned in \cite{132}, to extract the representative features from the fused data. It is also considered an area of methodology to keep the property of high accuracy and tackle obstacles of limited dataset size. After exploring the existing solutions on DBN and RBM, some of the current works focus on video tasks by applying these networks. The Deep Belief Network(DBN) architecture comprises the stacked Restricted Boltzmann Machine(RBMs) for analyzing three types of data: single-modality, multi-modality, and cross-modality which are mentioned in most research works. The entire structure mentioned in Figure 4 represents the structure of DBN and stacked RBM which are utilized in the action recognition tasks in the application of deep neural networks. The top of Figure 4 represents that the DBN(Deep Belief Network) is a structured deep neural network that is composed of stacked RBM, where the input neurons of each layer are the output neurons of the last layer. The final layers of RBM are followed by a multilayer perceptron(MLP) for outputting the behavior detection probability during multiple classification results. The bottom part of Figure 4 represents the neuron structures of each layer in RBM, where the output of neural networks is calculated by inputting the information through a hidden layer combined with weights and bias defined on each hidden unit and hidden layers between the input layer and output layer. 

\begin{figure}[!ht]
    \centering
     {\includegraphics[width=0.45\textwidth]{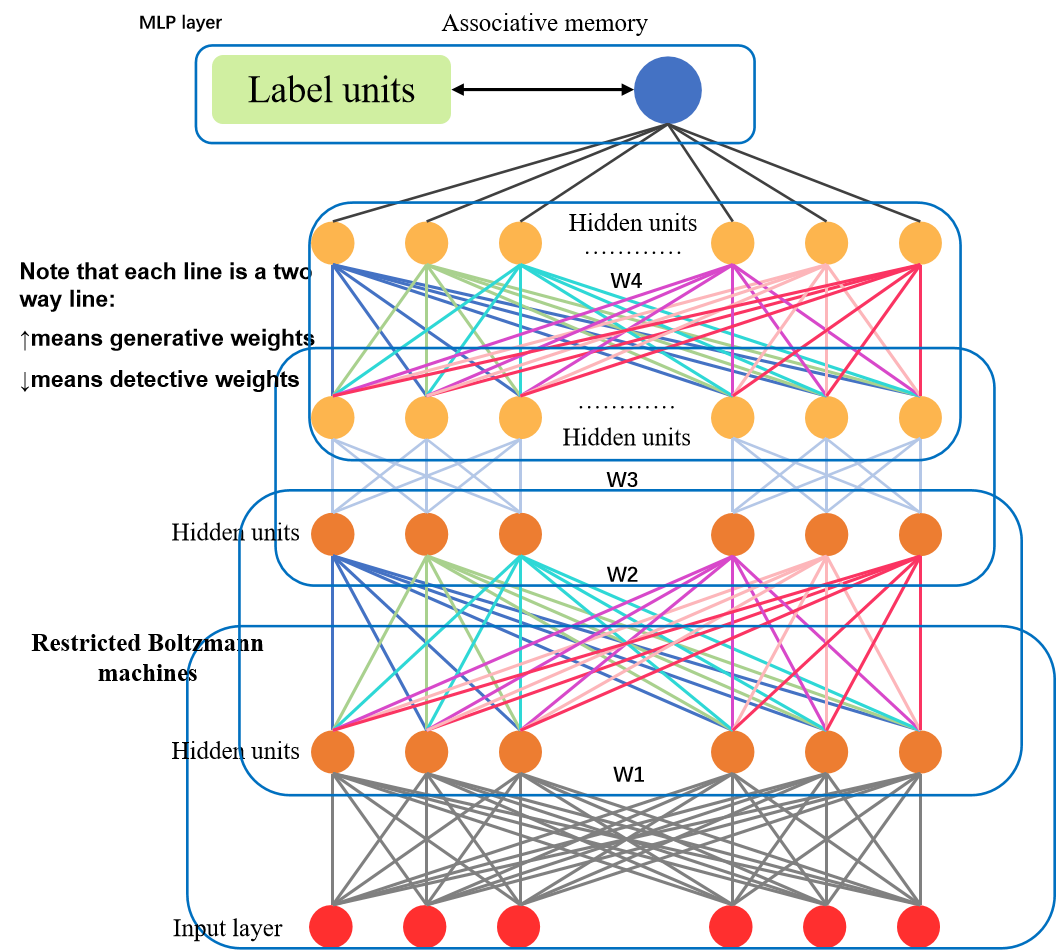}}
    \\
    {\includegraphics[width=0.37\textwidth]{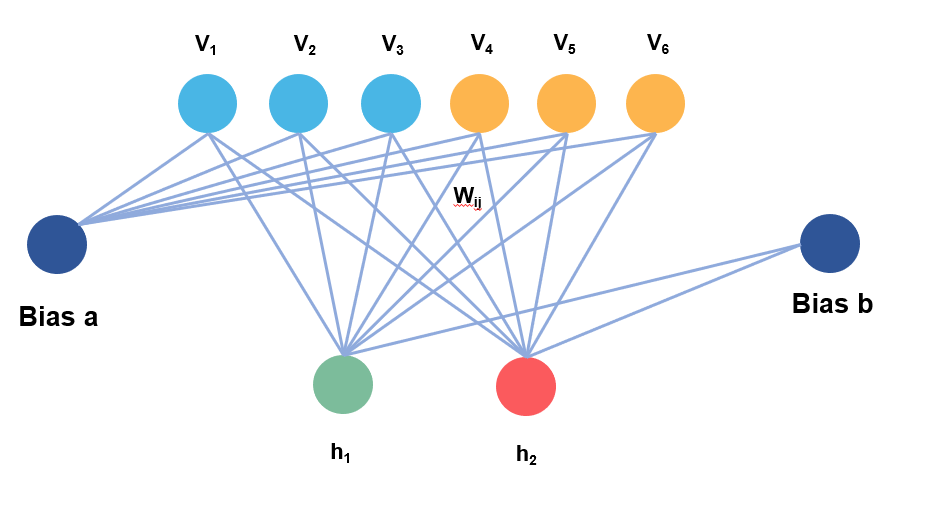}}
    \caption{Deep belief network for action recognition.}
    \label{fig:foobar}
\end{figure}

The research work \cite{133} compared the performance of the Deep Boltzmann Machine and Deep Belief Network, which said the Deep Boltzmann Machine cannot work individually when applying DBN models on large-scale datasets. The paper illustrated clearly the conceptions of Deep Belief Networks for activity recognition by defining stacked RBM to compose unsupervised DBN structure. The input images are also pre-processed by Gaussian transformation before inputting them into networks. The DBN structures, including stacked RBM for human action recognition, could be considered as following energy functions composing a size of \textit{m} * \textit{n} matrix called w{$_i$}{$_j$} of visible and hidden units，

\begin{equation}
\begin{pmatrix}
v_{11} & \cdots & v_{1m}\\
\vdots & \ddots &  \vdots \\
h_{n1} & \cdots & h_{mn}
\end{pmatrix} 
\end{equation}
where \textit{m} is the number of visible units and \textit{n} is the number of hidden units, \textit{i} and \textit{j} represents the index of hidden units and visible units, and b{$_i$} and b{$_j$} represent the offsets of visible neurons and hidden neurons respectively where the energy function is expressed as equation 2,  
\begin{equation}
\label{deqn_ex1}
E(v,h;\Theta) = -\sum_{i=1}^{m}b_{i}v_{i}-\sum_{j=1}^{n}b_{j}h_{j} - \sum_{i=1}^{m}*\sum_{j=1}^{n}v_{i}h_{j}w_{ij}
\end{equation}
and the value of hidden neurons and visible neurons are calculated according to Equation 3 and Equation 4,
\begin{equation}
\label{deqn_ex2}   
P(h_{j}) = P(h|v,\theta) = \sigma(b_{h_j} + \sum_{j=1}^{n}h_{j}w_{ij})
\end{equation}
\begin{equation}
\label{deqn_ex3}
P(v_{i}) = P(v|h, \theta) = \sigma(b_{v_i} + \sum_{i=1}^{m}w_{ij}v_i)
\end{equation}

where \textit{w}{$_i$}{$_j$} represents the weights allocated to the connections between hidden neurons and visible neurons. After the stack of RBMs, the DBN is ended by a multilayer perceptron being utilized for behavior classification result output. Where the MLP works as the foundation of multi-category classification with the structure of numerous inputs mapping to a single output result, and the production of MLP compares with a fixed threshold to determine whether it belongs to a specific category, which is represented as equation 5 and equation 6. 
\begin{equation}
\label{deqn_ex4}
 z = \sum_{k=1}^{m}w_{k}x_{k} + bias 
\end{equation}

\begin{equation}
\sigma(z) = \frac{1}{1 + e^{(-z)}}
\end{equation}
As mentioned above, the typical structure of DBN where multiple Restricted Boltzmann Machines are stacked, followed by the MLP layers, especially the proposed Convolution Deep Belief Network is to accomplish the human action recognition tasks. The research work presents DBN could achieve great classification accuracy compared with SVM. However, the limitation of this type of DBN is that it only recognizes single-modality activity rather than multi-modality activity. DBN has fewer variants focusing on various types of human actions when the accuracy is higher than other traditional methods. 

The analysis of multi-modal or the composition of activities requires a huge amount of data fusion, and the format of data varies. Due to the limitation of DBN on limited datasets, the survey on multi-modal referred to the process of applying the DBN on multiple types of input data, including image and text, where the word tokenizers must be considered during the processing \cite{132, 134}. With the usage of DBN-stacked RBM, the model can not only process the video data but also process the audio data by converting the audio information into tokenizers, especially the modality data, by fusing data as the category of late/early/hybrid. The hybrid model \cite{133} is also utilized for extracting information from multiple dimension video data, reaching the highest prediction accuracy during the comparisons of multiple schemes and deep neural networks. 

The DBNs have been worked as unsupervised classification methods for recognizing temporal and spatial information of videos. It tackled the issue of recognizing most videos without labels or partially labeled. Active learning made the action prediction accuracy lower than other types of deep neural networks when applying this methodology on top of unlabeled or partially labeled data. DBM and DBN solved this problem with the deep neural networks methodology \cite{135}. The performance of DBN architecture showed the recognition result had achieved high accuracy for action recognition through using Gibbs Sampling as an image pre-processing step. The application scenario of DBN also includes sports recognition. The proposed \cite{136} time-space deep belief network recognizes human sports behaviors, resulting in higher accuracy than CNN. Furthermore, contrastive divergence(CD) is the optimization algorithm of prediction accuracy using a weights matrix under DBN. 

\vspace{-1em}
\subsection{Graph-Like Methodologies}

The graph theory could also be applied to recognition tasks as some of the video data could be modeled into the structure of a graph, including graph convolutional networks, scene graphs, and graph neural networks. We analyzed some standard arrangements for classification and regression deep learning tasks using a graph-like design.

\subsubsection{Graph Convolutional Network}

The research works \cite{137,138} are composed of two graphs, one of which represents the body graph by using the matrix representing the bone’s relationship, where the relationships are represented as joints matrix. Furthermore, the temporal and spatial GCN is proposed for recognizing skeleton-based behaviors. RGB and RGB-D data are utilized during the research work to draw the skeleton-based human pose, Where spatial and temporal information is recognized by multiplying graph-like networks, which were investigated  and certified that the GCN and its variants have achieved excellent identification performance \cite{139}, especially on the skeleton-based analysis. The graph-based convolutional network is also utilized for temporal and spatial human skeleton-based analysis. The geometry information is also extracted from the original images(frames in the video). The extrinsic and intrinsic connections of physical parts are recognized as the relationship analysis of human behaviors. The dense block is connected in the above research works in temporal sequences. Similar to other graph-based methodologies, the input features are encoded with parameter matrix W which is the weights parameters of each neural layer and the outputs are the results from parameterize operations.

\begin{figure}[ht]
\centering
  \includegraphics[width=9.5cm, height=4cm]{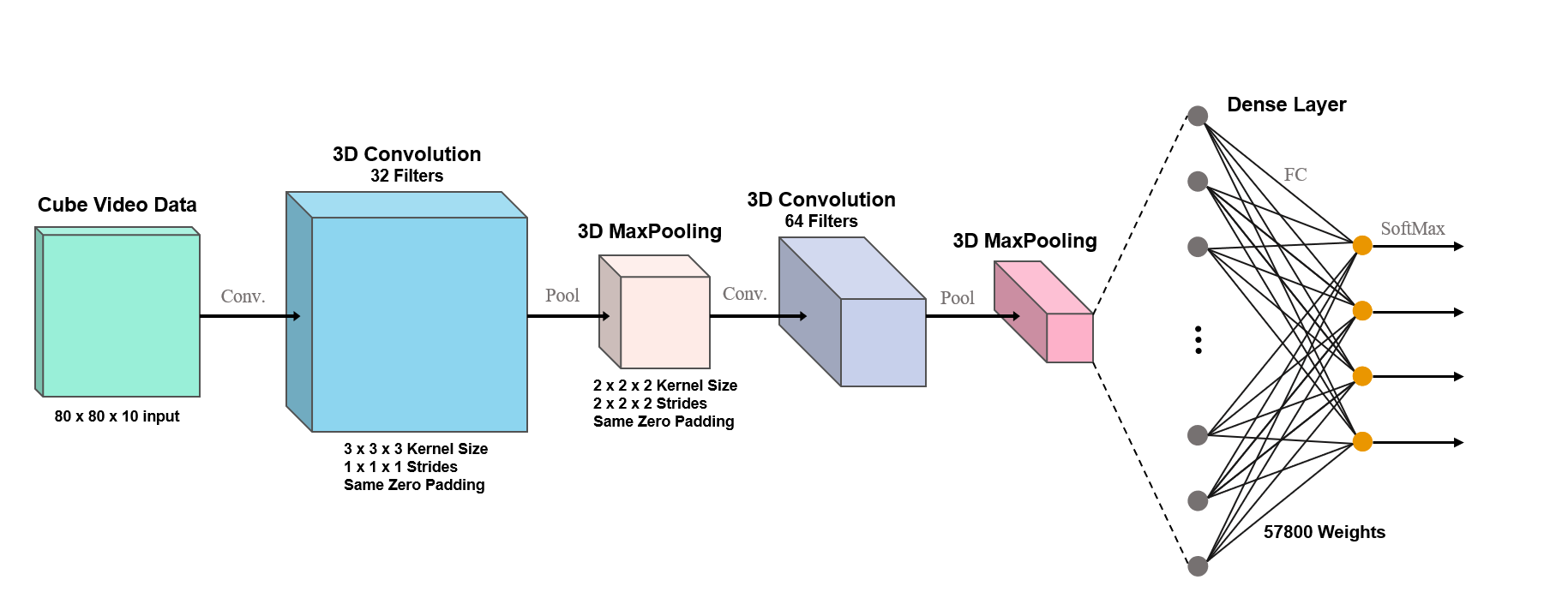}
  \caption{3D-ConvNet schemes.}
  \label{Figure}
\end{figure}

\subsubsection{Graph Neural Network}

The Graph Neural Network is also a type of neural network called MST-GNN framework for analyzing human skeleton body information based on NTU RGB+D datasets. MST-GNN is proposed to analyze the human skeleton graph, which combined the temporal and spatial elements of body information based on NTU RGB+D datasets \cite{140}. The GNN scheme proposed here combined computational graph units for temporal graph analysis \cite{140}. Data fusion is also utilized for both spatial and temporal feature extraction by utilizing the MST-GCU module, which worked as a part of the feature extractor.

\subsubsection{Graph Attention Networks}

The Graph Attention Network(GAT) utilized the aggregation-concatenation methods for the message-passing algorithms. Rather than Graph-based networks and Attention-based networks, the graph-attention neural network could work on the edge predictions and node predictions based on the graph structures. For the video-based human behavior recognition tasks, the Graph Attention Network could be manipulated on skeleton-based and spatial-temporal event recognitions, which include video behavior-level and pixel-level recognition tasks. Both temporal and spatial elements could be extracted as features from deep neural networks, where each layer represents a feature generator and learned patterns from defined inputs. As an individual network in a bunch of graph neural networks, such as graph interactive networks, graph convolutional networks and etc, the GAT works as the pioneer of graph-based transformers as a type of visual transformer.  

\section{Temporal Models on Behaviour Recognition Tasks}

\subsection{RNNs: Gated Recurrent Units(GRUs) and LSTM}
GRU is utilized for predicting the time sequential movements with multiple variants, including GRU-SVM and GRU-D models. The Gaussian and Kalman filters are the two most critical filters utilized for blurring images before the videos are inputted into networks. The input categories of videos could be split into two types, where one is the original video and another is processed by a Gaussian filter. Few works utilized individual Gated Recurrent Units for recognizing human actions. Therefore, the GRUs is nearly utilized individually instead of combined with other types of models. With the work \cite{lea2017temporal} utilizing the encoder-decoder structure for the representation of temporal feature extractions. This work proposed the methodology of a temporal convolutional network only to solve the long-time dependency problem rather than image features only. 

There are some works utilizing GRUs combined with CNN methodologies which proofed that the efficiency of hybrid modelings. Existing methods include the novel proposed LSTM and GRUs framework, which has reached a higher recognition accuracy which is higher than 90\%. From this perspective, we could see that the recognition tasks could utilize the GRUs structure with the update gate and reset gate \cite{143} for the recognition tasks. Furthermore, the gated recurrent unit could be utilized after the data comes from sensors. As the sensors' data are usually unstructured or semi-structured due to the application scenarios, such as health data, environment data, or climate change data which are largely impacted by outside aspects. 

Mostly, the RNNs represent a great range of time-sensitive tasks, such as frame-based behavior predictions on the basis of video sequences or behavior sequences. Or the abnormal behavior recognitions according to the previous normal behavior detection. From this perspective, the RNNs could work pretty well on the basis of context-based tasks. Usually, the LSTM and GRUs could work well not only on traditional image recognition tasks, but it could also work with attention-based tasks as a sub-kernel when it works as individual components. The GRUs are widely utilized in the area of signal processing, wifi communications, or wearable sensors in the wide usage of applications. 

\subsection{Vision Transformers and Adaptive Modules}
As widely known by plenty of works, due to the flexibility of the attention mechanism, transformers are widely utilized in lots of NLP tasks for information extraction. The vision transformer such as Vision Transformer(ViT) is a major transformer module designed for vision information detection. As the specific information required by video information extractions, the transformer structures are proven in a wide usage of behavior recognition, such as skeleton-based tasks \cite{tem2} or multi-modality tasks. In the research \cite{tem1} built two blocks with local and global temporal for the time-dependency feature extraction. The adaptive kernel learns the features from video clips frame by frame followed by the activation functions to determine the final output scores. Through the local and global attention modules, the attention weights are calculated through the attention modules. During the research work in \cite{tem1}, the change of backbone determined recognition accuracy from the frame dependency relationships. 

\begin{figure}[ht]
    \centering
  \includegraphics[scale=0.7]{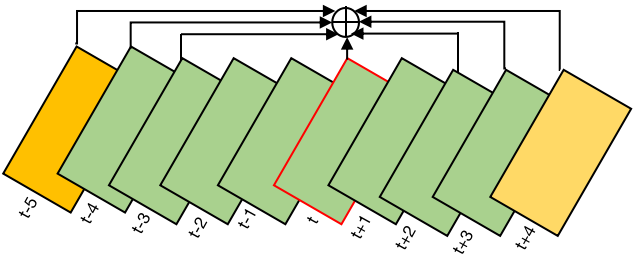}
  \caption{Temporal adaptive block of feature aggregation functions.}
  \label{Figure}
\end{figure}

The vision transformer has been proven efficient in multiple types of video-based tasks, especially in spatial-temporal task processing. The ViT could also be utilized in the tasks of deep representation learning tasks in cross-modal related tasks \cite{mmit}. Which is widely known by transformers, where the attention-based mechanism is applied to replace the convolution operation to accomplish the feature extraction in the behavior recognition tasks. There were multiple works proposed followed by \cite{tem1} for the adaptive block on the structure of video-based behavior recognition, which is a temporal feature extraction block. The adaptive block always works with the aggregation operator to aggregate the multiple temporal features from multiple frames as shown in Figure 6. All the work we illustrated here is to capture consistent behaviors through a series of video frames and extract the features from a single frame, then make the comparison during the change of modality or temporal location. So the aggregation function could work on multiple dimensions of input frames, such as pixel-level(spatial), frame-level(temporal), or object(detection) level.
\section{3D Convolutional Neural Networks and Hybrid Architectures}
\vspace{-0.5em}
\subsection{3D CNN Methodologies}
There are plenty of derived 3D Convolutional Neural Networks that have achieved outstanding feature extraction performance on both spatial and temporal through double-stream or multi-stream architectures. Currently, we have collected various types of Convolutional Neural Networks having multiple architectures with both hybrid and individual streams on behavior classification tasks. 

3D CNN with adaptive temporal feature resolutions specially outlined the temporal elements by using the bounding box with object detection(ROI, region of interests) through convolutional operation on pixel-wise video data to extract features on a temporal perspective \cite{144}. The procedure mentioned above is to get the similarity for mapping the temporal feature map to a low-level space. We verified that the ROI methodology could not help improve behavior recognition accuracy but helps reduce detected areas, which may be available on the computation cost deduction for behavior recognition. This procedure is still waiting to be verified in the research works. After we summarized most of the existing frameworks for a few past years, the hybrid models are more popular than the unique architecture for human action recognition tasks. Nevertheless, some examples include C3D embedded with RNNs for movement recognition on time-sequential tasks. Most of these frameworks are composed of multiple components to improve classification accuracy. 

During the proposed work \cite{145}, 3D-CNN is utilized to classify two types of human behaviors, including violent and non-violent, which could be utilized in surveillance videos for a series of video frames. The 3D-CNN is the stage of feature extractions, and the features are inputted by using attention-module to both spatial and temporal perspectives. As the property of terrorism events, the above recognition methodology classified the ordinarily polite and violent behaviors according to the temporal sequence of frames by the spatial and temporal order of 3D convolutional neural networks as the structure shown in Figure 5. The method mentioned above is combined with a Generative Adversarial Network(GAN) to separate abnormal and typical behaviors by defined constraints. 

\begin{figure}[ht]
  \includegraphics[width=9cm, height=4cm]{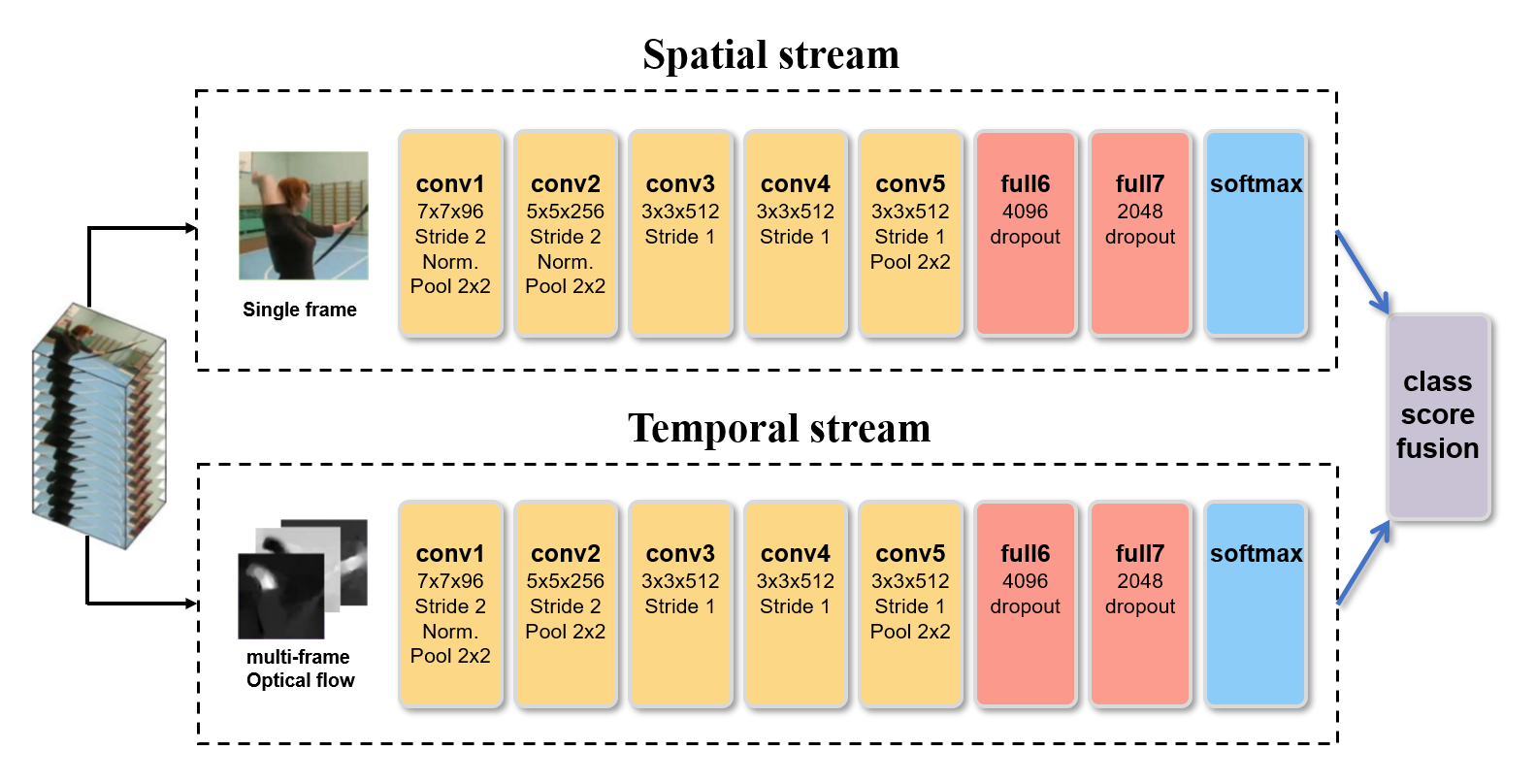}
  \caption{Double stream structure in spatial and temporal recognition.}
  \label{Figure}
\end{figure}

\subsection{3D-CNN and Hybrid Behavior Recognition Tasks}

\subsubsection{Multiple-stream Hybrid Architecture in 3D-CNN }
Multiple double-stream tasks could be discussed in our following articles. We collected a series of related works and summarized the following categories include a. optical-flow stream and RGB-stream, b. spatial and temporal streams, LSTM(time-dependency) streams, CNN streams, optical-flow streams, and soft-attention streams. Especially the multi-modality recognition, as the modality analysis required multiple types of data format, such as the combination of audio, text, and video. The fusion of multiple modalities will be helpful in the improvement of identification accuracy as the context of human recognition is critical under specific conditions. The research works \cite{126,127,128,146} mention how modality data fusion is fused for human behavior recognition as the stream structure could be separated into time dependency, which is the change on individual parts from \textit{T} to \textit{T} + ${\tau}$ as mentioned in above research work. The double stream \cite{126} includes two types of branches as shown in Figure 6. From the methods of fusion, data fusion methods could be separated into late fusion, early fusion, and hybrid fusion, especially in the perspective of temporal data and modality data fusion in double-stream or hybrid stream architectures. The first stream is a human skeleton branch with convolutional neural networks followed by an LSTM on time dependency analysis. And the second stream is composed of multiple branches, which include various views from different angles, as each angle will provide a different perspective and features. Finally, the outputs of multiple components combine and calculate the final prediction results. 

Besides the 3D convolutional neural networks, multiple types of hybrid architectures have implemented behavior recognition tasks, as mentioned in the article \cite{147}. During their research, they said how the double-stream architectures are utilized in spatial and temporal behavior recognition tasks. Meanwhile, they discussed the behavior prediction on multi-modality slow-fast models using multi-stream structures. Furthermore, hybrid architectures exist in generative adversarial networks(GAN) for group activity recognition and attention-based LSTM models \cite{126, 148}. A generator and a discriminator construct the generative adversarial network. The whole GAN pipeline extracted the features first and generated recognition scores through both the discriminator and the generator. Then GAN and graph-based networks could recognize the group activity through transfer learning or generative models. From this perspective, the recognition tasks are usually accomplished by combining LSTMs and other related models, especially the multiple streams and levels. The steps generally follow the steps combined by recognizing the bounding box, masking the objects by image segmentation, and finishing the classifications by the recognition procedure \cite{149}. During the works mentioned above, the article utilized the correlation matrix to measure the performance of classification results. In summary, the hybrid architecture extracted various features through individual branches and then finalized the classification score by utilizing fully connected layers to get the results of recognition and classification. 

The multiple streams will include the multi-channel processing of images, such as RGB, optical flow, and skeleton data. As mentioned in research \cite{150}, the human-robot interaction with deep neural networks detects behaviors by using the ResNet-101 and R3D-18 as backbones to see both spatial and temporal elements from both RGB and skeleton behavior patterns. The cross-subject and cross-view of this work have improved the performance on the behavior recognition tasks. But the above work has its limitation as it only utilized the 2D skeleton as input sequences with the standard methods to generate the probability of classification results. Similar to the above, the double stream attention-based LSTM methodology is provided in research work \cite{151}, with both spatial and temporal attention-aware for behavior recognition by extracting features from optical flow images. 

\subsubsection{Attention-Aware Hybrid Architecture in 3D-CNN}  
The attention-based hybrid end-to-end deep learning methodologies are also standard in recent years. We know that attention-based and attention-aware methods are well-known in other tasks, and lots of research work have verified that attention-aware processes are efficient in human and action recognition tasks. From this perspective, our review summarized existing research, such as in research work \cite{152, 153}, Lei. Arguments video data based on four types of the video stream and the modality data through 3D ResNet-50 as a backbone during the feature extraction process. They fused the results of prediction outputs and calculated the prediction
results into activation functions to simplify the output results. They utilized motion guided attention model to combine the development of 2 motion networks. It has reached better recognition results rather than others. As the modality data, such as videos and audio, could be utilized for modeling training, emotions/expressions, gestures, or gaits could also be extracted with the deep neural attention-based network. In research work \cite{154}, they built dual-stream models, including human action and visual attention trained based on UCF101 datasets. They made their backbone network for feature extraction and applied visual attention to behavior types, specifically facial expressions, and gestures. They utilized long-term engagement with the LSTM network on temporal dependency analysis. We mentioned that they considered spatial and temporal attention mechanisms with learnable parameters. The idea is to output the feature tensors through attention networks by replacing various
backbones of models, which achieved higher recognition performance on gestures and emotions recognition based on the human body parts.

Finally, we found that attention-based methodologies, especially soft attention, are effective in disorder recognition and correction. In research work \cite{155}, the soft-attention-based behavior recognition methodology is proposed for background and foreground operations extraction from video data. The results after attention steps are inputted into a pre-trained VGG model as the backbone to extract the human features from multiple views - front and side view. They also utilized the attention mechanism to store the generated driver behaviors segments from the original frames by fuzzing the backgrounds. And the hybrid model w/o backbones required multiple streams to work together and generate objective results rather than single-stream architectures.

\vspace{-0.4em}
\section{Remaining Obstacles in Action Recognition}

\subsection{Obstacles Solving in Spatial and Temporal}
One of the problems existing in current behavior identification tasks is splitting some confusing poses, such as climbing poles with pull-ups in the classification tasks of behavior recognition\cite{yang2022deep}. This problem is also found in the 3D skeleton pose estimation tasks. During the skeleton-based tasks as one of the modality data, the optical flow also has this type of problem for clearly extracting distinct behaviors from the high-similarity samples. As this problem first is addressed in facial expression recognition, then it was found in behavior recognition. Then the fusion of space and time is utilized for solving this type of problem. As consistent behavior has the property of the human part's location change, where long-range time analysis will heavily rely on the extracted temporal elements. We mentioned how the temporal elements are analyzed in the last section. While we also found that there are few works utilizing the background or context as features, most of the works only consider human motion pixels rather than background pixels which made the optical flow methods great fitting for temporal-related tasks.

The wide usage of end-to-end deep learning methodology\cite{yang2022deep, 14} is currently the most popular framework being utilized in behavior recognition tasks. But we found that there are few works to consider other types of architectures or software structures for spatial-temporal tasks, which may help in the improvement of training efficiency to reduce the computation costs and time spent. The vision Transformer broke and solved the situations from both space and time perspectives. The structure of the encoder-decoder also could accomplish the tasks of feature extractions with the downstream classification tasks. 

While the popularity of video-based human behavior recognition, the transformer architecture is also widely utilized in solving behavior recognition tasks. Rather than utilizing the deep neural networks as the backbone to accomplish the feature extraction tasks, the transformer-based methodologies are popular for identifying the temporal information, such as the frame-by-frame extraction or time-sensitive tasks through its attention mechanism, for example, swin transformer and vision transformer, especially in the usage of video-based tasks. The purpose of using a transformer on vision tasks is to solve the modality and temporal issues such as MM-Vit  \cite{chen2022mm}. To solve this type of problem, the vision transformer(ViT), video vision transformer(ViViT)\cite{vivit}, and video swin transformer are proposed to analyze 3D/4D information rather than the 2-D inputs only, the solution above is mainly converting the vision information into embeddings or tokens, but the usages or the applications of swin transformers on behavior recognitions are still limited. The solution is to stack multiple swin transformer blocks or ViT together for feature extraction on both temporal and modality perspectives. 
\vspace{-0.5em}
\subsection{Prediction of Futures and Usages}
The future of deep neural networks will be focusing on multi-modality such as images, videos, and sentences. During our investigations, deep neural networks not only focus on the usage of human behavior recognition but also be utilized in multiple disciplines, such as speech recognition. The application scenarios are switched from video recognition only to the multiple types of modality recognition. As the deep neural network has involved lots of layers for feature extractions, especially the pre-trained models such as ImageNet or ConceptNet for the conception generation based on the extracted features. Also with the wide usage of IoT devices, as well as edge computing, deep neural networks will be widely utilized in a wider range of situations for computation resources optimization as well as better feature extraction results for higher identification results generation even the auto modeling training. There are multiple perspectives that could be considered under the wide range of usages during deep neural networks with multiple scenarios in the application or theory impacts. 

\section{Conclusion And Further Work}

Firstly, our work investigated the existing recent and latest deep learning methodologies for human action recognition from both spatial and temporal perspectives. During the past several years of research, data types, including RGB, RGB-D, optical flow, grayscale, and skeleton-based, are all manipulated for video content analysis. Rather than the spatial information only, the video information provides a context for human action recognition, primarily when video recognition is utilized for abnormal detection and analysis when the actions depend on the environment. The temporal information provides the context of behaviors rather than single frames.  
Then we also provided our summary of existing works in both mathematics and technique views for video behavior recognition. The multiple types of architectures and modalities of data are mentioned in the usage of deep neural networks, including both individual and hybrid networks. Finally, the investigation work of our review has provided an overview of existing networks and datasets and the future of this area.

\section*{Acknowledgments}
Ideas/reconstruction of the article, article writing, and drafting are contributed provided by Prof. Y. Yang and Prof. Y. Li. The research work is financially and physically supported by the Lab of the School of Information Science and Engineering, Shandong University(Qingdao), China.

% $ biblatex auxiliary file $
% $ biblatex bbl format version 3.1 $
% Do not modify the above lines!
%
% This is an auxiliary file used by the 'biblatex' package.
% This file may safely be deleted. It will be recreated as
% required.
%
\printbibliography[title={References}]

\end{document}